# Reversible Jump MCMC Simulated Annealing for Neural Networks

Christophe Andrieu[‡] [*]   Nando de Freitas[†]   Arnaud Doucet[‡]

[‡]Engineering Department
Cambridge University
Cambridge CB2 1PZ, UK
{ad2,ca22}@eng.cam.ac.uk

[†] UC Berkeley Computer Science Division
387 Soda Hall, Berkeley
CA 94720-1776 USA
jfgf@cs.berkeley.edu

## Abstract

We propose a novel reversible jump Markov chain Monte Carlo (MCMC) simulated annealing algorithm to optimize radial basis function (RBF) networks. This algorithm enables us to maximize the joint posterior distribution of the network parameters and the number of basis functions. It performs a global search in the joint space of the parameters and number of parameters, thereby surmounting the problem of local minima. We also show that by calibrating a Bayesian model, we can obtain the classical AIC, BIC and MDL model selection criteria within a penalized likelihood framework. Finally, we show theoretically and empirically that the algorithm converges to the modes of the full posterior distribution in an efficient way.

## 1 INTRODUCTION

In this paper, we show that traditional model selection criteria within a penalized likelihood framework, such as Akaike's information criterion (AIC), minimum description length (MDL) and the Bayesian information criterion (BIC) (Akaike 1974, Schwarz 1985, Rissanen 1987), can be shown to correspond to particular hyper-parameter choices in a Bayesian formulation. That is, it is possible to calibrate the prior choices so that the problem of model selection within the penalized likelihood context can be mapped exactly to a problem of model selection via posterior probabilities. This technique has been previously applied in the areas of variable selection (George and Foster 1997) and the detection of harmonics in noisy signals (Andrieu and Doucet 1998).

After resolving the calibration problem, maximum likelihood estimation, with the aforementioned model selection criteria, is performed by maximizing the calibrated posterior distribution. To accomplish this goal, we propose an MCMC simulated annealing algorithm, which makes use of a homogeneous reversible jump MCMC kernel as proposal. This approach has the advantage that we can start with an arbitrary model order and the algorithm will perform dimension jumps until it finds the "true" model order. That is, one does not have to resort to the more expensive task of running a fixed dimension algorithm for each possible model order and subsequently selecting the best model. We also present a convergence theorem for the algorithm. The complexity of the problem does not allow for a comprehensive discussion in this short paper. Readers are encouraged to consult our technical report for further results and details (Andrieu, de Freitas and Doucet 1999)[1].

## 2 MODEL SPECIFICATION

We adopt the approximation scheme of Holmes and Mallick (1998), consisting of a mixture of $k$ RBFs and a linear regression term. (The work can, however, be straightforwardly extended to many other interesting inference and learning problems, such as fMRI time series modeling, wavelet networks, multivariate adaptive regression splines (MARS), Bayesian networks, etc.) This model is given by:

$$\mathbf{y}_t = \mathbf{b} + \boldsymbol{\beta}'\mathbf{x}_t + \mathbf{n}_t \quad k = 0$$

$$\mathbf{y}_t = \sum_{j=1}^{k} \mathbf{a}_j \phi(\|\mathbf{x}_t - \boldsymbol{\mu}_j\|) + \mathbf{b} + \boldsymbol{\beta}'\mathbf{x}_t + \mathbf{n}_t \quad k \geq 1$$

where $\|\cdot\|$ denotes a distance metric (usually Euclidean or Mahalanobis), $\boldsymbol{\mu}_j \in \mathbb{R}^d$ denotes the $j$-th RBF centre for a model with $k$ RBFs, $\mathbf{a}_j \in \mathbb{R}^c$ denotes the $j$-th RBF amplitude and $\boldsymbol{b} \in \mathbb{R}^c$ and $\boldsymbol{\beta} \in \mathbb{R}^d \times \mathbb{R}^c$ denote the linear regression parameters. The noise sequence $\mathbf{n}_t \in \mathbb{R}^c$ is assumed to be zero-mean white Gaussian. It is important to mention that although the dependency of $b$, $\beta$ and $\mathbf{n}_t$ on $k$ has not been made explicit, these parameters are indeed affected by the value of $k$.

---

[*]Authorship based on alphabetical order.

[1]The software is available at the following website http://www.cs.berkeley.edu/~jfgf.

Depending on our *a priori* knowledge about the smoothness of the mapping, we can choose different types of basis functions (Girosi, Jones and Poggio 1995). The most common choices are linear, cubic, thin plate spline and Gaussian. For convenience, the approximation model is expressed in vector-matrix form[2]:

$$\mathbf{y} = \mathbf{D}(\boldsymbol{\mu}_{1:k,1:d}, \mathbf{x}_{1:N,1:d})\boldsymbol{\alpha}_{1:1+d+k,1:c} + \mathbf{n}_t$$

That is:

$$\mathbf{y} = \begin{bmatrix} y_{1,1} \cdots y_{1,c} \\ y_{2,1} \cdots y_{2,c} \\ \vdots \\ y_{N,1} \cdots y_{N,c} \end{bmatrix} \quad \boldsymbol{\alpha} = \begin{bmatrix} b_1 \cdots b_c \\ \beta_{1,1} \cdots \beta_{1,c} \\ \vdots \\ \beta_{d,1} \cdots \beta_{d,c} \\ \mathbf{a}_{1,1} \cdots \mathbf{a}_{1,c} \\ \vdots \\ \mathbf{a}_{k,1} \cdots \mathbf{a}_{k,c} \end{bmatrix}$$

$$\mathbf{D} = \begin{bmatrix} 1 & x_{1,1} \cdots x_{1,d} & \phi(\mathbf{x}_1, \boldsymbol{\mu}_1) \cdots \phi(\mathbf{x}_1, \boldsymbol{\mu}_k) \\ 1 & x_{2,1} \cdots x_{2,d} & \phi(\mathbf{x}_2, \boldsymbol{\mu}_1) \cdots \phi(\mathbf{x}_2, \boldsymbol{\mu}_k) \\ \vdots & \vdots & \vdots \\ 1 & x_{N,1} \cdots x_{N,d} & \phi(\mathbf{x}_N, \boldsymbol{\mu}_1) \cdots \phi(\mathbf{x}_N, \boldsymbol{\mu}_k) \end{bmatrix}$$

where the noise process is assumed to be normally distributed $\mathbf{n}_t \sim \mathcal{N}(0, \sigma_i^2)$ for $i = 1, \ldots, c$. It should be stressed that $\sigma^2$ depends implicitly on the model order $k$. The number $k$ of RBFs and their parameters $\boldsymbol{\theta} \triangleq \{\boldsymbol{\alpha}_{1:m,1:c}, \boldsymbol{\mu}_{1:k,1:d}, \sigma_{1:c}^2\}$, with $m = 1 + d + k$, are unknown. Given the data set $\{\mathbf{x}, \mathbf{y}\}$, the objective is to estimate $k$ and $\boldsymbol{\theta} \in \Theta_k$.

## 3 PROBABILISTIC MODEL

In (Andrieu, de Freitas and Doucet 1999, Andrieu, de Freitas and Doucet 2000), we follow a Bayesian

---

[2]The notation $\mathbf{y}_{1:N,1:c}$ is used to denote an $N$ by $c$ matrix, where $N$ is the number of data and $c$ the number of outputs. That is, $\mathbf{y}_{1:N,j} \triangleq (y_{1,j}, y_{2,j}, \ldots, y_{N,j})'$ denotes all the observations corresponding to the $j$-th output ($j$-th column of $\mathbf{y}$). To simplify the notation, $\mathbf{y}_t$ is equivalent to $\mathbf{y}_{t,1:c}$. That is, if one index does not appear, it is implied that we are referring to all of its possible values. Similarly, $\mathbf{y}$ is equivalent to $\mathbf{y}_{1:N,1:c}$. The shorter notation will be favored, while the longer notation will be invoked to avoid ambiguities and emphasize certain dependencies. This notation, although complex, is essential to avoid ambiguities in the design of the reversible jump algorithm.

approach where the unknowns $k$ and $\boldsymbol{\theta}$ are regarded as being drawn from appropriate prior distributions. These priors reflect our degree of belief in the relevant values of these quantities (Bernardo and Smith 1994). An hierarchical prior structure is used to treat the priors' parameters (hyper-parameters) as random variables drawn from suitable distributions (hyper-priors).

Here, we focus on performing model selection using classical criteria such AIC, BIC and MDL. We show that performing model selection using these criteria is equivalent to computing the joint maximum a posteriori (MAP) of a "calibrated" posterior distribution. This interpretation allows one to develop very efficient simulated annealing algorithms to solve this difficult global optimization problem.

### 3.1 PENALIZED LIKELIHOOD MODEL SELECTION

Traditionally, penalized likelihood model order selection strategies, based on standard information criteria, require the evaluation of the maximum likelihood (ML) estimates for each model order. The number of required evaluations can be prohibitively expensive unless appropriate heuristics are available. Subsequently, a particular model $\mathcal{M}_s$ is selected if it is the one that minimizes the sum of the log-likelihood and a penalty term that depends on the model dimension (Djurić 1998, Gelfand and Dey 1997). In mathematical terms, this estimate is given by:

$$\mathcal{M}_s = \underset{\mathcal{M}_k : k \in \{0, \ldots, k_{\max}\}}{\arg\min} \left\{ -\log(p(\mathbf{y}|k, \widehat{\boldsymbol{\theta}}, \mathbf{x})) + \mathcal{P} \right\} \quad (1)$$

where $\widehat{\boldsymbol{\theta}} = (\widehat{\boldsymbol{\alpha}}_{1:m}, \widehat{\boldsymbol{\mu}}_{1:k}, \widehat{\sigma}_k^2)$ is the ML estimate of $\boldsymbol{\theta}$ for model $\mathcal{M}_k$. $\mathcal{P}$ is a penalty term that depends on the model order. Examples of ML penalties include the well known AIC, BIC and MDL information criteria (Akaike 1974, Schwarz 1985, Rissanen 1987). The expressions for these in the case of Gaussian observation noise are:

$$\mathcal{P}_{\text{AIC}} = \xi \quad \text{and} \quad \mathcal{P}_{\text{BIC}} = \mathcal{P}_{\text{MDL}} = \frac{\xi}{2} \log(N)$$

where $\xi$ denotes the number of model parameters ($k(c+1) + c(1+d)$ in the case of an RBF network). These criteria are motivated by different factors: AIC is based on expected information, BIC is an asymptotic Bayes factor and MDL involves evaluating the minimum information required to transmit some data and a model, which describes the data, over a communications channel.

Using the conventional estimate of the variance for

Gaussian distributions:

$$\hat{\sigma}_i^2 = \frac{1}{N}(\mathbf{y}_{1:N,i} - \mathbf{D}(\hat{\boldsymbol{\mu}}_{1:k},\mathbf{x})\hat{\boldsymbol{\alpha}}_{1:m,i})'$$
$$\times (\mathbf{y}_{1:N,i} - \mathbf{D}(\hat{\boldsymbol{\mu}}_{1:k},\mathbf{x})\hat{\boldsymbol{\alpha}}_{1:m,i})$$
$$= \frac{1}{N}\mathbf{y}_{1:N,i}'\mathbf{P}_{i,k}\mathbf{y}_{1:N,i}$$

where $\mathbf{P}_{i,k}$ is the least squares orthogonal projection matrix:

$$\mathbf{I}_N - \mathbf{D}(\hat{\boldsymbol{\mu}}_{1:k},\mathbf{x})\left[\mathbf{D}'(\hat{\boldsymbol{\mu}}_{1:k},\mathbf{x})\mathbf{D}(\hat{\boldsymbol{\mu}}_{1:k},\mathbf{x})\right]^{-1}\mathbf{D}'(\hat{\boldsymbol{\mu}}_{1:k},\mathbf{x})$$

we can re-express equation (1) as follows:

$$\mathcal{M}_s = \arg\max_{\mathcal{M}_k:k\in\{0,\ldots,k_{\max}\}} \left\{ \left[\prod_{i=1}^c (\mathbf{y}_{1:N,i}'\mathbf{P}_{i,k}\mathbf{y}_{1:N,i})^{-N/2}\right] \times \exp(-\mathcal{P}) \right\} \quad (2)$$

### 3.2 BAYESIAN MODEL

We place the following uninformative prior on the parameters:

$$p(\boldsymbol{\alpha}_{1:m},\boldsymbol{\mu}_{1:k},\sigma^2|k) = \prod_{i=1}^c \mathbb{I}_\Omega(k,\boldsymbol{\mu}_{1:k})\frac{(\mathbf{D}'\mathbf{D})^{1/2}}{\sigma_i^2}$$

where $\Omega$ denotes the joint space of $k$ and $\boldsymbol{\mu}$ and $\mathbb{I}$ is the standard set indicator variable. This prior is represented by the Bayesian network of Figure 1.

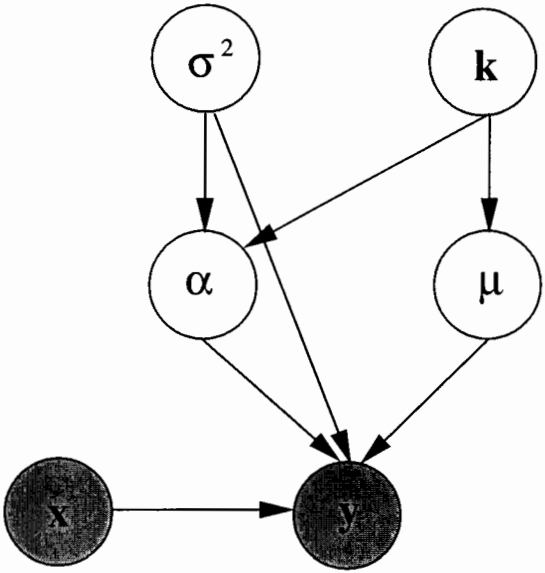

Figure 1: Directed acyclic graph for the Bayesian prior.

For this prior it is possible to integrate out the variance and coefficients to obtain the following expression for the marginal posterior distribution $p(k,\boldsymbol{\mu}_{1:k}|\mathbf{x},\mathbf{y})$:

$$\propto \left[\prod_{i=1}^c \left(\mathbf{y}_{1:N,i}'\mathbf{P}_{i,k}\mathbf{y}_{1:N,i}\right)^{(-N/2)}\right]\mathbb{I}_\Omega(k,\boldsymbol{\mu}_k)p(k) \quad (3)$$

Let us now define the MAP estimate of this distribution as follows:

$$(k,\boldsymbol{\mu}_{1:k})_{\text{MAP}} = \arg\max_{k,\boldsymbol{\mu}_{1:k}\in\Omega} p(k,\boldsymbol{\mu}_{1:k}|\mathbf{x},\mathbf{y}) \quad (4)$$

Comparing equations (2) and (3), we note that these expressions agree whenever:

$$p(k) \propto \exp(-\mathcal{P}) \propto \exp(-\mathcal{C}k)$$

This proportionality ensures that the expression for the calibrated posterior $p(k,\boldsymbol{\mu}_{1:k}|\mathbf{x},\mathbf{y})$ corresponds to the term that needs to be maximized in the penalized likelihood framework (equation (2)). Note that for the purposes of optimization, we only need the proportionality condition with $\mathcal{C} = c+1$ for the AIC criterion and $\mathcal{C} = (c+1)\log(N)/2$ for the MDL and BIC criteria.

It has thus been shown that by calibrating the priors in the Bayesian formulation, one can obtain the expression that needs to be maximized in the classical penalized likelihood formulation with AIC, MDL and BIC model selection criteria. Consequently, the penalized likelihood framework can be interpreted as a problem of maximizing the joint posterior posterior distribution $p(k,\boldsymbol{\mu}_{1:k}|\mathbf{x},\mathbf{y})$.

The sufficient conditions that need to be satisfied so that the distribution $p(k,\boldsymbol{\mu}_{1:k}|\mathbf{x},\mathbf{y})$ is proper are not overly restrictive. Firstly, $\Omega$ has to be a compact set, which is not a problem in our setting. Secondly, $\mathbf{y}_{1:N,i}'\mathbf{P}_{i,k}\mathbf{y}_{1:N,i}$ has to be larger than zero for $i = 1,\ldots,c$. In (Andrieu, de Freitas and Doucet 1999), it is shown that this is the case unless $\mathbf{y}_{1:N,i}$ spans the space of the columns of $\mathbf{D}(\boldsymbol{\mu}_{1:k},\mathbf{x})$, in which case $\mathbf{y}_{1:N,i}'\mathbf{P}_{i,k}\mathbf{y}_{1:N,i} = 0$. This event has zero measure.

## 4 REVERSIBLE JUMP SIMULATED ANNEALING

We can solve the stochastic optimization problem posed in the previous subsection by using a simulated annealing strategy. The simulated annealing method (Geman and Geman 1984, Van Laarhoven and Arts 1987) involves simulating a non-homogeneous Markov chain whose invariant distribution at iteration $i$ is no longer equal to $\pi(\mathbf{z})$, but to:

$$\pi_i(\mathbf{z}) \propto \pi^{1/T_i}(\mathbf{z})$$

where $T_i$ is a decreasing cooling schedule with $\lim_{i\to+\infty} T_i = 0$. The reason for doing this is that,

under weak regularity assumptions on $\pi(\mathbf{z})$, $\pi^\infty(\mathbf{z})$ is a probability density that concentrates itself on the set of global maxima of $\pi(\mathbf{z})$.

The simulated annealing method with distribution $\pi(\mathbf{z})$ and proposal distribution $q(\mathbf{z}^\star|\mathbf{z})$ involves sampling a candidate value $\mathbf{z}^\star$ given the current value $\mathbf{z}$ according to $q(\mathbf{z}^\star|\mathbf{z})$. The Markov chain moves towards $\mathbf{z}^\star$ with probability $\mathcal{A}_{\text{SA}}(\mathbf{z}, \mathbf{z}^\star) = \min\left\{1, \left(\pi^{1/T_i}(\mathbf{z}) q(\mathbf{z}^\star|\mathbf{z})\right)^{-1} \pi^{1/T_i}(\mathbf{z}^\star) q(\mathbf{z}|\mathbf{z}^\star)\right\}$, otherwise it remains equal to $\mathbf{z}$.

To obtain an efficient algorithm it is of paramount importance to choose and efficient proposal distribution. If we choose a homogeneous transition kernel $\mathcal{K}(\mathbf{z}, \mathbf{z}^\star)$ that satisfies the following reversibility property:

$$\pi(\mathbf{z}^\star)\mathcal{K}(\mathbf{z}^\star, \mathbf{z}) = \pi(\mathbf{z})\mathcal{K}(\mathbf{z}, \mathbf{z}^\star)$$

it follows that:

$$\mathcal{A}_{\text{SA}} = \min\left\{1, \frac{\pi^{(1/T_i - 1)}(\mathbf{z}^\star)}{\pi^{(1/T_i - 1)}(\mathbf{z})}\right\} \quad (5)$$

We propose to use as transition kernel a reversible jump MCMC algorithm (Green 1995). This is a general state-space Metropolis-Hastings (MH) algorithm (see (Andrieu, Djurić and Doucet 1999) for an introduction). One proposes candidates according to a set of proposal distributions. These candidates are randomly accepted according to an acceptance ratio which ensures reversibility and thus invariance of the Markov chain with respect to the posterior distribution. Here, the chain must move across subspaces of different dimensions, and therefore the proposal distributions are more complex: see (Green 1995, Richardson and Green 1997) for details. For our problem, the following moves have been selected:

1. Birth of a new basis by proposing its location in an interval surrounding the input data.

2. Death of an existing basis by removing it at random.

3. Merge a randomly chosen basis function and its closest neighbor into a single basis function.

4. Split a randomly chosen basis function into two neighbor basis functions, such that the distance between them is shorter than the distance between the proposed basis function and any other existing basis function. This distance constraint ensures reversibility.

5. Update the RBF centres.

These moves are defined by heuristic considerations, the only condition to be fulfilled being to maintain the correct invariant distribution. A particular choice will only have influence on the convergence rate of the algorithm. The birth and death moves allow the network to grow from $k$ to $k+1$ and decrease from $k$ to $k-1$ respectively. The split and merge moves also perform dimension changes from $k$ to $k+1$ and $k$ to $k-1$. The merge move serves to avoid the problem of placing too many basis functions in the same neighborhood. On the other hand, the split move is useful in regions of the data where there are close components. Other moves may be proposed, but the ones suggested here have been found to lead to satisfactory results.

The resulting transition kernel of the simulated Markov chain is then a mixture of the different transition kernels associated with the moves described above. This means that at each iteration one of the candidate moves, birth, death, merge, split or update, is randomly chosen. The probabilities for choosing these moves are $b_k$, $d_k$, $m_k$, $s_k$ and $u_k$ respectively, such that $b_k + d_k + m_k + s_k + u_k = 1$ for all $0 \leq k \leq k_{\max}$. A move is performed if the algorithm accepts it. For $k = 0$ the death, split and merge moves are impossible, so that $d_0 \triangleq 0$, $s_0 \triangleq 0$ and $m_0 \triangleq 0$. The merge move is also not permitted for $k = 1$, that is $m_1 \triangleq 0$. For $k = k_{\max}$, the birth and split moves are not allowed and therefore $b_{k_{\max}} \triangleq 0$ and $s_{k_{\max}} \triangleq 0$.

Consequently, the following algorithm, with $b_k = d_k = m_k = s_k = u_k = 0.2$, can find the joint MAP estimate of $\boldsymbol{\mu}_{1:k}$ and $k$:

---

**Reversible Jump Simulated Annealing**

1. Initialization: set $\left(k^{(0)}, \theta^{(0)}\right) \in \Theta$.

2. Iteration $i$.

   - Sample $u \sim \mathcal{U}_{[0,1]}$ and set the temperature with a cooling schedule.
   - If $(u \leq b_{k^{(i)}})$
     - then "birth" move (See Section 3.2.2).
     - else if $(u \leq b_{k^{(i)}} + d_{k^{(i)}})$ then "death" move (See Section 3.2.2).
     - else if $(u \leq b_{k^{(i)}} + d_{k^{(i)}} + s_{k^{(i)}})$ then "split" move (See Section 3.2.3).
     - else if $(u \leq b_{k^{(i)}} + d_{k^{(i)}} + s_{k^{(i)}} + m_{k^{(i)}})$ then "merge" move (See Section 3.2.3).
     - else update the RBF centres (See Section 3.2.1).
     
     End If.
   - Perform an MH step with the annealed acceptance ratio (equation (5)).

3. $i \leftarrow i + 1$ and go to 2.

4. Compute the coefficients $\alpha_{1:m}$ by least squares (optimal in this case):

$$\widehat{\alpha}_{1:m,i} = [\mathbf{D}'(\boldsymbol{\mu}_{1:k},\mathbf{x})\mathbf{D}(\boldsymbol{\mu}_{1:k},\mathbf{x})]^{-1}\mathbf{D}'(\boldsymbol{\mu}_{1:k},\mathbf{x})\mathbf{y}_{1:N,i}$$

∎

In the algorithm, we fixed the mixing mixture coefficients $b_k$, $d_k$, $m_k$ and $s_k$ to 0.2. The simulated annealing moves are explained in the following subsections.

**Remark 1** *Note that our algorithm has allowed us to integrate out the coefficients $\alpha_{1:m}$. It can thus be stated that the variance of this Rao Blackwellised estimate is less than the variance that would have resulted if we had sampled $\alpha_{1:m}$ jointly with $\mu$ and $k$ (Casella and Robert 1996, Liu, Wong and Kong 1994).*

### 4.0.1 Update Move

The radial basis centres are sampled one-at-a-time using an MH algorithm. To accomplish this, we use a mixture of random walk proposals and a global proposal surrounding the input data: see (Andrieu, de Freitas and Doucet 1999) for details. The resulting target distribution $p(\boldsymbol{\mu}_{j,1:d}|\mathbf{x},\mathbf{y},\boldsymbol{\mu}_{-j,1:d})$ is proportional to:

$$\left[\prod_{i=1}^{c}(\mathbf{y}'_{1:N,i}\mathbf{P}_{i,k}\mathbf{y}_{1:N,i})^{(-\frac{N}{2})}\right]\exp(-\mathcal{P})$$

and, consequently, the acceptance ratio $\mathcal{A}_{\text{RJSA}}$ is given by:

$$\min\left\{1,\left[\prod_{i=1}^{c}(\frac{\mathbf{y}'_{1:N,i}\mathbf{P}_{i,k}\mathbf{y}_{1:N,i}}{\mathbf{y}'_{1:N,i}\mathbf{P}^{\star}_{i,k}\mathbf{y}_{1:N,i}})^{(\frac{N}{2})}\right]\right\}$$

where $\mathbf{P}^{\star}_{i,k}$ is similar to $\mathbf{P}_{i,k}$ with $\boldsymbol{\mu}_{1:k,1:d}$ replaced by $\{\boldsymbol{\mu}_{1,1:d},\boldsymbol{\mu}_{2,1:d},\dots,\boldsymbol{\mu}_{j-1,1:d},\boldsymbol{\mu}^{\star}_{j,1:d},\boldsymbol{\mu}_{j+1,1:d},\dots,\boldsymbol{\mu}_{k,1:d}\}$.

### 4.0.2 Birth and Death Moves

Suppose that the current state of the Markov chain is in $\{k\}\times\Theta_k$, then the birth and death moves are given by:

---

**Birth move**

- Propose a new RBF centre at random from the space surrounding x.

- Evaluate $\mathcal{A}_{birth}$, see equation(7), and sample $u\sim\mathcal{U}_{[0,1]}$

- If $u\leq\mathcal{A}_{birth}$ then the state of the Markov chain becomes $(k+1,\boldsymbol{\mu}_{1:k+1})$, else it remains equal to $(k,\boldsymbol{\mu}_{1:k})$.

---

**Death move**

- Choose the basis centre, to be deleted, at random among the $k$ existing bases.

- Evaluate $\mathcal{A}_{death}$, see equation (7), and sample $u\sim\mathcal{U}_{[0,1]}$

- If $u\leq\mathcal{A}_{death}$ then the state of the Markov chain becomes $(k-1,\boldsymbol{\mu}_{1:k-1})$, else it remains equal to $(k,\boldsymbol{\mu}_{1:k})$.

∎

---

The acceptance ratio for the proposed birth move is deduced from the following expression (Green 1995):

$$r_{birth} \triangleq (posterior\ distributions\ ratio)$$
$$\times (proposal\ ratio) \times (Jacobian) \quad (6)$$

That is:

$$r_{birth} = \left[\prod_{i=1}^{c}(\frac{\mathbf{y}'_{1:N,i}\mathbf{P}_{i,k}\mathbf{y}_{1:N,i}}{\mathbf{y}'_{1:N,i}\mathbf{P}_{i,k+1}\mathbf{y}_{1:N,i}})^{(\frac{N}{2})}\right]\frac{\Im\exp(-\mathcal{C})}{k+1}$$

The Jacobian in this case is equal to 1. Similarly,

$$r_{death} = \left[\prod_{i=1}^{c}(\frac{\mathbf{y}'_{1:N,i}\mathbf{P}_{i,k}\mathbf{y}_{1:N,i}}{\mathbf{y}'_{1:N,i}\mathbf{P}_{i,k-1}\mathbf{y}_{1:N,i}})^{(\frac{N}{2})}\right]\frac{k\exp(\mathcal{C})}{\Im}$$

Hence, the acceptance probabilities corresponding to the described moves are:

$$\mathcal{A}_{birth}=\min\{1,r_{birth}\},\ \mathcal{A}_{death}=\min\{1,r_{death}\} \quad (7)$$

### 4.0.3 Split and Merge Moves

The merge move involves randomly selecting a basis function $(\boldsymbol{\mu}_1)$ and then combining it with its closest neighbor $(\boldsymbol{\mu}_2)$ into a single basis function $\boldsymbol{\mu}$, whose new location is:

$$\boldsymbol{\mu}=\frac{\boldsymbol{\mu}_1+\boldsymbol{\mu}_2}{2}$$

The corresponding split move that guarantees reversibility is:

$$\begin{aligned}\boldsymbol{\mu}_1 &= \boldsymbol{\mu}-u_{ms}\varsigma^{\star}\\ \boldsymbol{\mu}_2 &= \boldsymbol{\mu}+u_{ms}\varsigma^{\star}\end{aligned}\quad(8)$$

where $\varsigma^{\star}$ is a simulation parameter and $u_{ms}\sim\mathcal{U}_{[0,1]}$. Note that to ensure reversibility, the merge move is only performed if $\|\boldsymbol{\mu}_1-\boldsymbol{\mu}_2\|<2\varsigma^{\star}$. Suppose now that the current state of the Markov chain is in $\{k\}\times\Theta_k$, then:

---

**Split move**

- Randomly choose an existing RBF centre.

- Substitute it for two neighbor basis functions, whose centres are obtained using equation (8). The distance (typically Euclidean) between the new bases has to be shorter than the distance between the proposed basis function and any other existing basis function.

- Evaluate $\mathcal{A}_{split}$, see equation(9), and sample $u \sim \mathcal{U}_{[0,1]}$

- If $u \leq \mathcal{A}_{split}$ then the state of the Markov chain becomes $(k+1, \mu_{1:k+1})$, else it remains equal to $(k, \mu_{1:k})$.

**Merge move**

- Choose a basis centre at random among the $k$ existing bases. Then find the closest basis function to it applying some distance metric, e.g. Euclidean.

- If $\|\mu_1 - \mu_2\| < 2\varsigma^*$, substitute the two basis functions for a single basis function in accordance with equation (4.0.3).

- Evaluate $\mathcal{A}_{merge}$, see equation (9), and sample $u \sim \mathcal{U}_{[0,1]}$.

- If $u \leq \mathcal{A}_{merge}$ then the state of the Markov chain becomes $(k-1, \mu_{1:k-1})$, else it remains equal to $(k, \mu_{1:k})$. ∎

---

The acceptance ratios for the proposed split and merge moves are given by:

$$r_{split} = \left[\prod_{i=1}^{c} \left(\frac{\mathbf{y}'_{1:N,i}\mathbf{P}_{i,k}\mathbf{y}_{1:N,i}}{\mathbf{y}'_{1:N,i}\mathbf{P}_{i,k+1}\mathbf{y}_{1:N,i}}\right)^{(\frac{N}{2})}\right] \frac{k\varsigma^* \exp(-\mathcal{C})}{k+1}$$

and

$$r_{merge} = \left[\prod_{i=1}^{c} \left(\frac{\mathbf{y}'_{1:N,i}\mathbf{P}_{i,k}\mathbf{y}_{1:N,i}}{\mathbf{y}'_{1:N,i}\mathbf{P}_{i,k-1}\mathbf{y}_{1:N,i}}\right)^{(\frac{N}{2})}\right] \frac{k \exp(\mathcal{C})}{\varsigma^*(k-1)}$$

The Jacobian corresponding to the split move is equal to:

$$J_{split} = \left|\frac{\partial(\mu_1, \mu_2)}{\partial(\mu, u_{ms})}\right| = \begin{vmatrix} 1 & 1 \\ -\varsigma^* & \varsigma^* \end{vmatrix} = 2\varsigma^*$$

The acceptance probabilities for these moves are:

$$\mathcal{A}_{split} = \min\{1, r_{split}\}, \; \mathcal{A}_{merge} = \min\{1, r_{merge}\} \quad (9)$$

## 5 CONVERGENCE

We have the following convergence theorem:

**Theorem 1** *Under certain assumptions found in (Andrieu, Breyer and Doucet 1999), the series of $(\boldsymbol{\theta}^{(i)}, k^{(i)})$ converges in probability to the set of global maxima, that is for a constant $\epsilon > 0$, it follows that:*

$$\lim_{i \to \infty} \Pr\left(\frac{\pi(\boldsymbol{\theta}^{(i)}, k^{(i)})}{(\boldsymbol{\theta}_{\max}, k_{\max})} \geq 1 - \epsilon\right) = 1$$

***Proof.*** *In (Andrieu, de Freitas and Doucet 1999), we show that the transition kernels for each temperature are uniformly geometrically ergodic. Hence, as a corollary of (Andrieu, Breyer and Doucet 1999, Theorem 1), the convergence result for the simulated annealing MCMC algorithm follows* ∎

## 6 EXPERIMENT: ROBOT ARM MAPPING

The robot arm data set is often used as a benchmark to compare neural network algorithms[3]. It involves implementing a model to map the joint angle of a robot arm $(x_1, x_2)$ to the position of the end of the arm $(y_1, y_2)$. The data were generated from:

$$y_1 = 2.0\cos(x_1) + 1.3\cos(x_1 + x_2) + \epsilon_1$$
$$y_2 = 2.0\sin(x_1) + 1.3\sin(x_1 + x_2) + \epsilon_2$$

where $\epsilon_i \sim \mathcal{N}(0, \sigma^2)$; $\sigma = 0.05$. We use the first 200 observations of the data set to train our models and the last 200 observations to test them. In the simulations, we selected cubic basis functions.

We tested the reversible jump simulated annealing algorithms with the AIC and MDL criteria on this problem. The results for the MDL criterion are depicted in Figure 2. We note that the posterior increases stochastically with the number of iterations and, eventually, converges to a maximum. The figure also illustrates the convergence of the train and test set errors for each network in the Markov chain. For the final network, we chose the one that maximized the posterior. This network consisted of 12 basis functions and incurred a mean-square error of 0.00512 in the test set. Following the same procedure, the AIC network consisted of 27 basis functions and incurred an error of 0.00520 in the test set. Our mean square errors are of the same magnitude as the ones reported by other researchers (Holmes and Mallick 1998, Mackay 1992, Neal 1996, Rios Insua and Müller 1998); slightly better. Our results in (Andrieu, de Freitas and Doucet 1999) indicate that the full Bayesian hierarchical model provides slightly more accurate results. The Monte Carlo integrations are, however, much more computationally demanding than the stochastic optimization task. They can take take up to 500000 iterations, whether the algorithm discussed here only required 200 iterations to obtain a reasonable solution.

---

[3] The robot arm data set can be found at http://wol.ra.phy.cam.ac.uk/mackay/

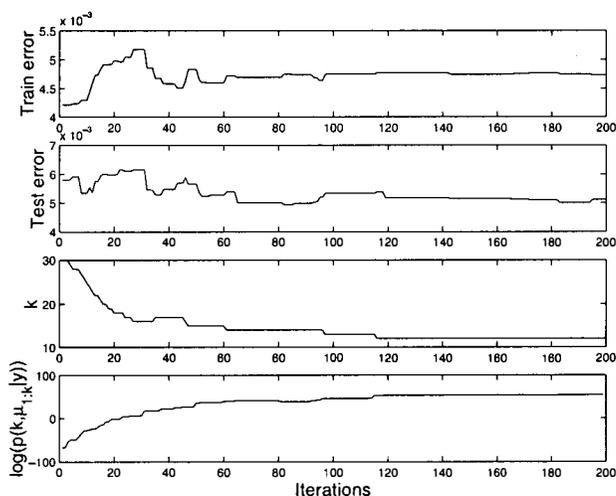

Figure 2: Performance of the reversible jump simulated annealing algorithm for 200 iterations on the robot arm data, with the MDL criterion.

## 7  CONCLUSIONS

We presented an efficient MCMC stochastic optimization algorithm that performs parameter estimation and model selection simultaneously. We also showed that starting from a full hierarchical Bayesian prior for neural networks, it is possible to derive the classical AIC, BIC and MDL penalized likelihood model selection criteria. Finally, we presented a convergence proof for the algorithm and showed, by means of an experiment, that the method produces very accurate results.